# LSTM-based Load Forecasting Robustness Against Noise Injection Attack in Microgrid

Amirhossein Nazeri[1], Pierluigi Pisu[1]

[1]Clemson University International Center for Automotive Research (CU-ICAR), SC, USA

*Abstract*— In this paper, we investigate the robustness of an LSTM neural network against noise injection attacks for electric load forecasting in an ideal microgrid. The performance of the LSTM model is investigated under a black-box Gaussian noise attack with different SNRs. It is assumed that attackers have just access to the input data of the LSTM model. The results show that the noise attack affects the performance of the LSTM model. The load prediction means absolute error (MAE) is 0.047 MW for a healthy prediction, while this value increases up to 0.097 MW for a Gaussian noise insertion with SNR= 6 dB. To robustify the LSTM model against noise attack, a low-pass filter with optimal cut-off frequency is applied at the model's input to remove the noise attack. The filter performs better in case of noise with lower SNR and is less promising for small noises.

I. INTRODUCTION

Forecasting electrical load has always been crucial for the utilities. To maintain the equilibrium between supply and demand, an accurate load prediction is required. Recurrent neural networks (RNNs) often outperform DNNs and CNNs when dealing with time-series because they have memory unit that stores valuable past data to aid in future prediction [1]. Despite this, RNN has issues with disappearing and expanding gradients as well as forgetting long-term dependencies in long-term data. To tackle this problem, Long Short-Term Memory (LSTM) networks are introduced [2]. Due to its numerous advantages, such as energy efficiency, improved stability, and resiliency, microgrids are becoming more common. Wireless measurements, online data transfer, and communication across units increase the efficiency and controllability of microgrids, but it makes them susceptible to cyber security risks including False Data Injection (FDI). Although machine learning models perform well for classification and regression problems, they are not robust enough to handle unexpected inputs. This model can perform significantly worse when exposed to cyberattacks and chaotic real-world input data.

In this paper we deliberately add Gaussian noise to the input of model, to investigate its severity on model's performance. Finally, we design and add a simple low-pass filter with optimal cut-off frequency to alleviate the impact of noise injection attack on LSTM model prediction.

## II. DEEP LSTM-BASED LOAD FORECASTING MODEL

The LSTM model makes predictions for the next five hours (60 data points) based on the prior four days (1152 data points). The input layer, two hidden LSTM layers, two Dropout layers, a fully connected feed-forward dense layer are the components of the model architecture. The LSTM model is implemented using TensorFlow 2.9 and Python 3.9.12 libraries. Two hidden layers with 128 LSTM units each make up the model, along with a dense layer with 60 neurons, two dropout regularization layers placed after each hidden layer. The proposed deep LSTM model's architecture is shown in Fig. 1, and the trained model's performance on the test set is shown in Fig. 2. In Fig. 1, M= 1152, N= 1152, P= 128, Q= 60.

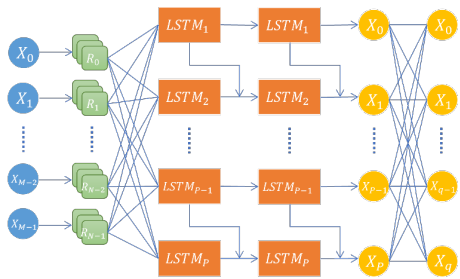

Figure 1: The architecture of the proposed deep LSTM model.

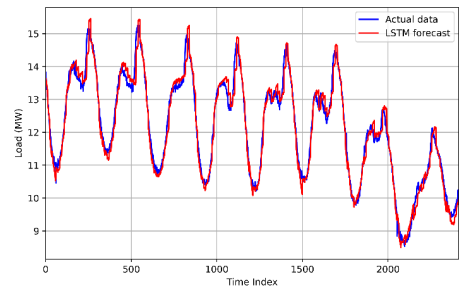

Figure 2: The forecasted load of the trained deep LSTM model on the test set.

## III. NOISE INJECTION ATTACK IN LOAD FORECASTING

This section investigates the impact of noise attack, to the input of the LSTM load forecaster. An ideal microgrid is considered for this purpose to simplify the track of noise injection flow and shed light just on the load forecasting unit performance. In the ideal case we assume that the load output of the microgrid is the same as historical data of the same day of last year. Also, the microgrid controllers are able to stabilize the frequency no matter what the predicted load initial points are.

The microgrid updates the LSTM load forecaster every 5 hours with observed measured load values. The measurement includes 60 datapoints and is used to produce the next batch of input data for LSTM future's prediction. Fig 3 shows the schematic of the generalized idea for this paper. $P_L^t$ and $\hat{P}_L^{t+1}$ correspond to power load and predicted power load at time $t$ and $t+1$, respectively. We assumed that the noise injection attack occurs after LSTM input being updated by measured data from microgrid and right before being inputted for prediction.

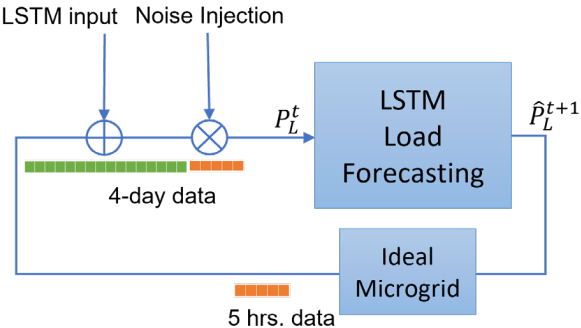

Figure 3: Noise injection attack in an ideal microgrid loop.

IV. NOISE INJECTION ATTACK

In this study, the noise is considered to be a gaussian noise profile with average value of zero and different SNR values. Noises with SNR values of 6, 10, 13, 20 dBs are constructed to examine the robustness of a typical multivariate time-series LSTM neural network. We do not consider the SNR range below 6dB because this range is easily detectable by the system anomaly detector. The noises profile along with some input load data for LSTM is depicted in Fig 4.

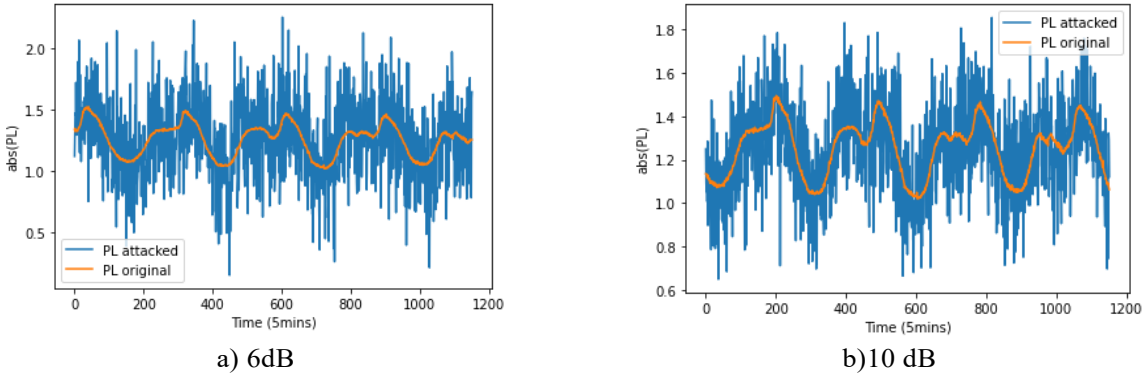

a) 6dB    b) 10 dB

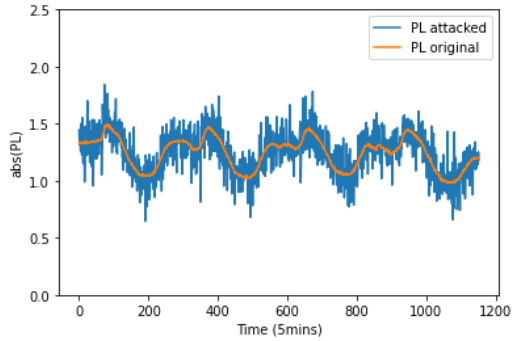 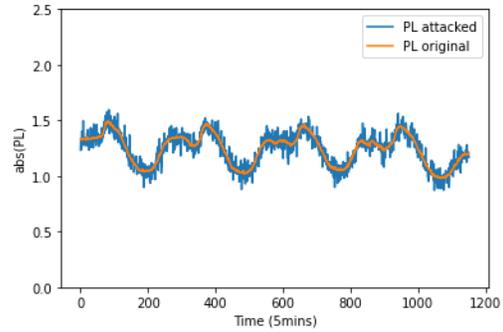

        c) 13 dB                                           d) 20 dB

Figure 4: different noise profiles for SNRs: 6, 10, 13, 20 dB.

Noise profiles in Fig. 4, are representatives of all possible noise injection attacks for this dataset. Noise profiles with SNR out of 6dB-20dB is not of interest due to the nature of load demand data. It can either be detected easily by anomaly detectors (<6dB SNR) or it leaves a negligible impact on load prediction (>20dB SNR). Table 1 shows the LSTM model prediction' mean absolute error (MAE) comparison between LSTM with no-attack and under noise injection attack scenario. As Table 1 indicates there is a huge rise of MAE value by reduction of SNR. Noise in input data with SNR= 6dB increases the model prediction MAE amount by a factor more than two times compared to the original prediction MAE.

Table 1. LSTM model prediction MAE for noise with different SNRs.

| SNR(dB) | MAE (MW) |
|---|---|
| No noise | 0.047 |
| 20 | 0.0619 |
| 13 | 0.0698 |
| 10 | 0.082 |
| 6 | 0.097 |
| Avg. MAE for SNRs 6-20dB | **0.079** |

Table 2. LSTM model prediction MAE for denoiser with different SNRs.

| SNR(dB) | MAE (MW) |
|---|---|
| - | - |
| 20 | 0.069 |
| 13 | 0.072 |
| 10 | 0.074 |
| 6 | 0.073 |
| Avg. MAE for SNRs 6-20dB | **0.073** |

## V. NOISE ATTACK REMOVAL

In this section, we remove the unknown injected noise attack from the LSTM input. To remove the noise attack, we first convert the input data into the frequency domain by taking the Fast Fourier Transform (FFT). Then we investigate the noise FFT coefficients in the attacked signal frequency spectrum. Fig 5. Shows the FFT results of the original and the attacked input signals with different SNRs.

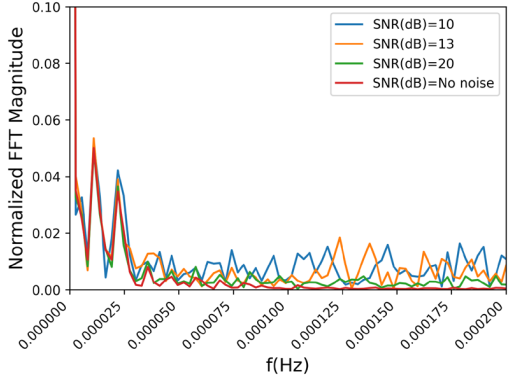 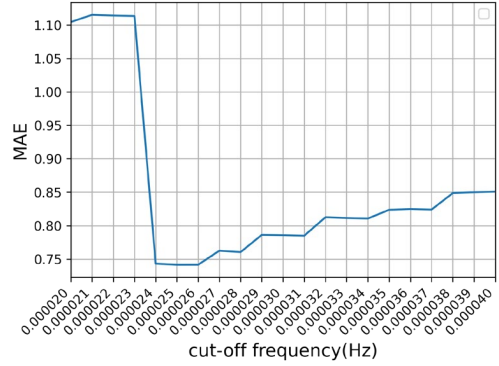

Figure 5: FFT analysis of healthy and attacked LSTM inputs

Figure 6: The low-pass filter performance versus cut-off frequency

According to Fig. 5, two frequencies 1.157e-5 Hz and 2.315e-5 Hz are the two dominant frequencies of both healthy and attacked load signals. The characteristic coefficients are affected by noise attack at higher frequencies. To remove the unknown noise, we implement a low-pass filter at the beginning of the LSTM unit before the input layer. We employ Grid-search optimization to achieve the optimal cut-off frequency. (1) shows the Grid-search optimization algorithm to find the optimal cut-off frequency. The optimal cut-off frequency minimizes the sum of absolute errors (SAEs) for different SNRs, between attacked signal and the reconstructed signal. Because the SNR value is unknown in real attack scenario, we include all possible SNRs, ranging 6-20dB, into the objective function.

Minimize:

$$\sum_{j=6}^{20} \sum_{i=1}^{N} \sum_{t=1}^{T} \left| P_{L_{Original}}^{t,i,j} - P_{L_{Reconstructed}}^{t,i,j}(f) \right| \quad , \quad f \in \{0.000020\text{Hz}, 0.000021\text{Hz}, \ldots, 0.000040\text{Hz}\} \quad (1)$$

$f$ is cut-off frequency for the noise cancellation filter from 0.000020Hz – 0.000040Hz. $j$ is noise SNR in dB that ranges from 6-20 dB. N=1750 times-series are used (one-year load data) each including T= 1152 records (obtained every 5 minutes). Fig. 6 demonstrates the results of the Grid-search for optimal cut-off frequency. According to Fig. 6, the low-pass filter with a cut-off frequency of 2.5e-5Hz presents the best performance in minimizing the original signal reconstruction error. Table 2 shows the model enhancement using the low pass filter. From Table 1 and 2 we conclude that the average MAEs is reduced by using the denoising filter. However, the filter performs better at lower SNRs when the signal undergoes a stronger noise attack, and it performs less promising at higher SNRs.

## V. CONCLUSIONS AND FUTURE WORKS

In this paper, we presented an LSTM model for load forecasting and investigated its performance against noise attacks. We also designed a simple low-pass filter to eliminate the noise. However, the low-pass filter performs better at lower SNRs and reduces the prediction performance at higher SNRs. In the next step, we will introduce the concept of smart auto-encoders(AEs) as a robust denoising mechanism to remove the noise insertion by learning from the noisy attack input and classifying the noise based on its SNR.